# A Deep Dive into the Design Space of a Dynamically Reconfigurable Cryogenic Spiking Neuron

Md Mazharul Islam[1], Shamiul Alam[1], *Graduate Student Member, IEEE,* Catherine D Schuman[1], Md Shafayat Hossain[2], Ahmedullah Aziz[1], *Member, IEEE*

**Abstract-Spiking neural network offers the most bio-realistic approach to mimic the parallelism and compactness of the human brain. A spiking neuron is the central component of an SNN which generates information-encoded spikes. We present a comprehensive design space analysis of the superconducting memristor (SM)-based electrically reconfigurable cryogenic neuron. A superconducting nanowire (SNW) connected in parallel with an SM function as a dual-frequency oscillator and two of these oscillators can be coupled to design a dynamically tunable spiking neuron. The same neuron topology was previously proposed where a fixed resistance was used in parallel with the SNW. Replacing the fixed resistance with the SM provides an additional tuning knob with four distinct combinations of SM resistances, which improves the reconfigurability by up to ~70%. Utilizing an external bias current ($I_{bias}$), the spike frequency can be modulated up to ~3.5 times. Two distinct spike amplitudes (~1V and ~1.8 V) are also achieved. Here, we perform a systematic sensitivity analysis and show that the reconfigurability can be further tuned by choosing a higher input current strength. By performing a 500-point Monte Carlo variation analysis, we find that the spike amplitude is more variation robust than spike frequency and the variation robustness can be further improved by choosing a higher $I_{bias}$. Our study provides valuable insights for further exploration of materials and circuit level modification of the neuron that will be useful for system-level incorporation of the neuron circuit.**

## I. INTRODUCTION

The persistent downsizing of transistors throughout the last several decades has given rise to the humongous technological revolution in all spheres of life. However, the continuation of this technological progression is no longer possible as we have approached the fundamental physical bottleneck [1], [2]. Besides, we are also hindered by the architecture-level limitation of separate memory and processing block known as the von Neumann bottleneck [3]. These limitations are impeding us to cope with the ever-increasing data growth and technological advancement has become more challenging [4], [5]. Researchers are trying to breach these limitations from all different directions through both hardware and software level improvement [6], [7]. In this context, neuromorphic computing has emerged as one of the most promising alternative computational paradigms to overcome these challenges by mimicking a brain-like in-memory architecture [8], [9]. Software-level implementation of neuromorphic computing offers suboptimal performances and suffers from complex architecture and large area overhead and thus fails to comply with its primary objective [10]. To achieve optimal performance and brain-like compactness, researchers have driven their efforts to build dedicated neuromorphic hardware in a wide variety of hardware platforms [11]. TrueNorth and Loihi are some of the noteworthy neuromorphic hardware implementations which offer significant computational merit with complete architectural features [12]–[14]. However, these implementations are still far from the level of energy efficiency and compactness of the human brain [15]. In this regard, researchers are constantly exploring novel hardware paradigms that will make the dreams of making energy-efficient brain-like machines possible. Among the emerging hardware paradigms, cryogenic electronics show great promise in terms of speed and energy requirement[16]–[21]. Cryogenic electronics have already been a center of focus for several years due to the emergence of the quantum computer and the immense interest in Rapid Single Flux Quantum circuits [22]. The ultra-high speed and ultra-low switching energy of cryogenic devices are unprecedented in any other hardware platforms that operate at the room temperature[23]. This inspired the researchers to explore neuromorphic hardware based on cryogenic devices which led to a unique and exciting paradigm- cryogenic neuromorphic hardware [24]. Josephson junction (JJ), quantum phase slip junction (QPSJ), and superconducting nanowire (SNW) are some of the superconducting devices that have been utilized to design neuromorphic circuits with superior performance [25]–[31]. Cryogenic devices have brought several significant advantages to the neuromorphic domain: (i) high intrinsic switching speed (several ps) (ii) ultra-low energy dissipation (~atto-Joules), (iii) dissipation-less interconnect, etc. Among the various topologies, Crotty *et al.* proposed an energy-efficient JJ-based spiking neuron [32]. The same topology was used by Toomey *et al.* to utilize an SNW-based spiking neuron with a higher fanout capability and a much simpler fabrication constraint [29]. However, this design contains a single tuning knob for reconfiguring the spiking rate, which is the bias current. This is practically infeasible as accurate biasing is practically challenging and offers only volatile reconfigurability [30]. A recent modification of the same neuron topology has incorporated a superconducting

M. M. Islam, Shamiul Alam, Catherine D Schuman and A. Aziz are with the Department of Electrical Engineering and Computer Science, University of Tennessee, Knoxville, TN, USA. (E-mail: mislam49@vols.utk.edu, salam10@vols.utk.edu , cschuman@utk.edu, aziz@ utk.edu, aziz@utk.edu).

Md Shafayat Hossain is with the Department of Physics, Princeton UNiversity, Princeton, NJ, USA. (E-mail: mdsh@princeton.edu)

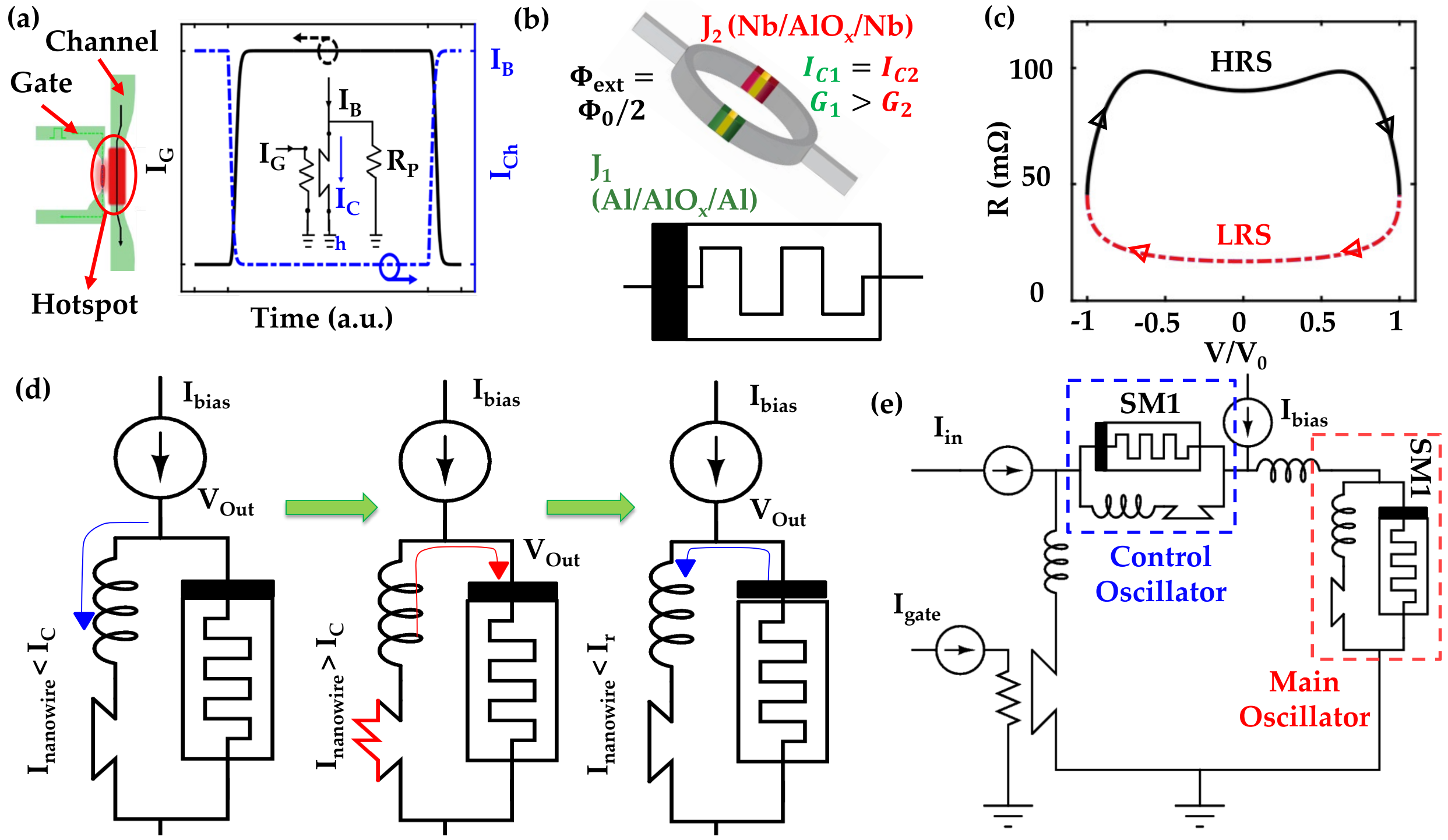

**Fig. 1: (a)** Schematic and switching characteristics of an *h-Tron* [35]. The channel current ($I_{Ch}$) is bypassed to the parallel resistance ($R_P$), when the gate current ($I_G$) is applied. **(b)** Structure and symbol of an SM [34]. The structure is formed by two JJs with phase-dependent conductance asymmetry **(c)** R-V characteristics of the SM displaying the two resistive states (LRS and HRS) [23]. **(d)** Structure and dynamics of the SM-based oscillator structure [33]. **(e)** SM-based neuron topology with two coupled oscillators [33].

memristor (SM) in the design to achieve a significantly higher (~75% higher) spike reconfigurability. The SM-based neuron offers a broader range of spike-rate with lower energy dissipation thanks to the milliohms ($m\Omega$s) range of non-volatile resistance levels. However, since the SM is primarily a JJ-based structure, our proposed design has a lower fanout capability compared to the solely nanowire-based design proposed by Toomey *et al*. Furthermore, the proposed design's fabrication simplicity is compromised due to the utilization of a Josephson junction-based structure, which introduces a higher degree of complexity in the fabrication process. Even so, our proposed SM-based design provides more reliable and distinct synaptic strength mapping, despite the reduced fanout capability. The neuronal dynamics of the proposed SM-based design are primarily governed by the unique interplay between the SNW and SM. Therefore, a systematic design space exploration is vital to identify the material/device-circuit co-design opportunities of the neuron and its further incorporation at the system level.

In this work, we perform a comprehensive design-space evaluation of the SM-based neuron topology. We have considered all the possible circuit variables for our design space exploration. We have varied the nanowire current parameters (critical current $I_c$, re-trapping current $I_r$), nanowire inductance, and the resistance levels of superconducting memristors to evaluate the operation sensitivity of the circuit components. We further extend our analysis by imposing random variation on the device parameters and evaluating the variation tolerance of our neuron topology. The unique contributions of this paper are as follows.

Table I: Materials and device simulation parameters

| Materials and Device Parameters | | | |
|---|---|---|---|
| SNW parameters | | SM parameters | |
| $R_{NW}$ | 5 kΩ | $R_{LRS}$ | 14.4 mΩ |
| $L_{NW}$ | 10 nH | $R_{HRS}$ | 98 mΩ |
| $I_c$, $I_r$ | 30 $\mu$A, 20 $\mu$A | $\gamma_0$ | 60° |

1) We perform sensitivity analysis of the neuronal spikes on the materials and geometric parameters of SNW.
2) We evaluate the levels of spiking rate and amplitude on the resistance levels of SM.
3) To verify the variation tolerance of our design, we perform rigorous variation analysis through Monte Carlo simulation.

## II. Superconducting Nanowire-based Reconfigurable Neuron

We begin our discussion by briefly describing our previously proposed neuron topology and its dynamics [33]. An SNW-based oscillator lies at the core of our design. An SNW shunted with a resistance generates oscillatory voltage under appropriate current bias ($I_{bias}$). When $I_{bias} > I_c$, the SNW turns resistive, and the current is redirected to $R_S$ giving rise to the voltage. Meanwhile, the current across the SNW reduces below $I_r$ and voltage is redirected towards SNW causing the fall of output voltage. These two cycles continue to generate oscillation at the output. Here, the oscillation period is primarily governed by the nanowire inductance ($L_{NW}$) and the shunted resistance ($R_s$). Hence, with a fixed $R_{NW}$ and $L_{NW}$, the oscillator generates a single frequency oscillation at the output. Toomey *et al.* proposed a neuron topology by incorporating two of these oscillators. In our previously proposed reconfigurable neuron topology, we have modified this oscillator structure by replacing the fixed resistance with a dual resistive superconducting structure referred to as a superconducting memristor (SM). This device was originally proposed by Peotta *et al* (Figs. 1(b,c)) [34].This SM structure utilizes the phase-dependent conductance of the superconducting tunnel junction (STJ) to build a conductance asymmetric SQUID (CA-SQUID) (Figs. 1(b,c)) [34]. It exhibits a non-volatile pinched hysteresis in its current-voltage ($I$–$V$) characteristics reminiscent of a conventional memristor. The rise time of the oscillation voltage is determined by the time constant $\tau_1 = L_{NW}/(R_s + R_{hs})$, where $R_s$ is the SM resistance, $R_{hs}$ is the resistance of the nanowire (non-SC mode), and $L_{NW}$ denotes the nanowire inductance. The fall time is proportional to the fall time constant, $\tau_2 = L_{NW}/R_s$. Thus, the oscillation period is proportional to the overall time constant ($\tau = \tau_1 + \tau_2$). By programming the SM in its LRS or HRS, it is possible to dynamically tune the oscillation frequency (Figs. 1(c,d)). The two oscillation frequencies (145 MHz and 82.6 MHz) correspond to the LRS and HRS of SM [33].

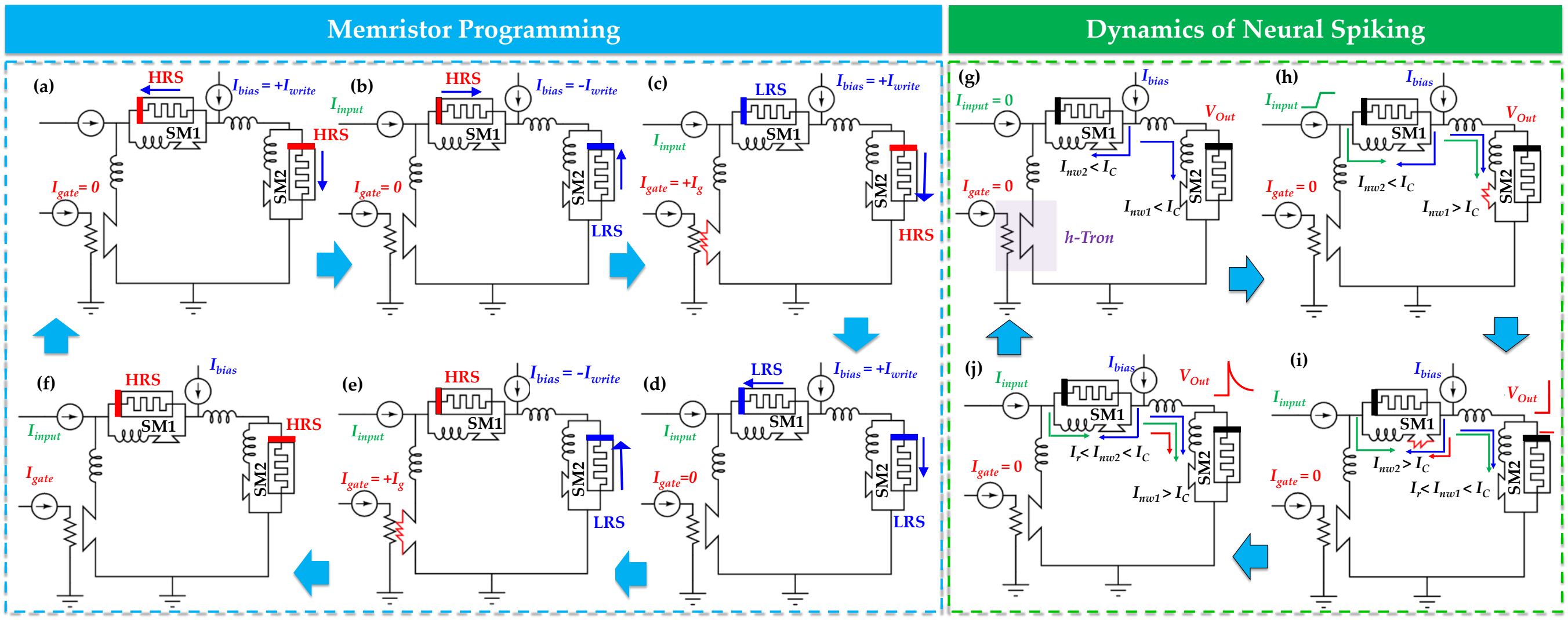

**Fig. 2: (a)-(f)** Different stages of the programming dynamics of SM-based neuron [33]. **(g)-(j)** Neural Spiking dynamics of the SM-based neuron [33].

Leveraging this feature of our oscillator, we designed a coupled oscillator-based reconfigurable spiking neuron topology (Fig. 1(e)). In addition, we have used a superconducting switching device, namely the heater cryotron (*h*-Tron) (Fig. 1(a)) [33], [35]. The *h*-Tron selectively directs the bias current depending on $I_{gate}$. In our design, the *h*-Tron provides individual programmability of the SMs (SM1 and SM2). Incorporation of the dual resistive SM in our design has improved the reconfigurability (The ratio of maximum and minimum achievable spiking rate) of our design by ~70%.

## III. Spike-Rate and Spike-Amplitude Reconfigurability

The complete programming scheme is shown in Figs. 2(a-f). To achieve full programmability, the *h*-Tron gate current ($I_{gate}$) and bias current ($I_{bias}$) are utilized in a combined fashion. When no gate current flws across the *h*-Tron gate ($I_{gate} = 0$), $I_{\text{bias}}$ is divided equally between the control oscillator and the main oscillator. With a sufficient strength of $I_{bias}$, the resistive states of both SM1 and SM2 can be programmed (Figs. 1(c,e)). However, when $I_{gate}$ is applied, the channel of the *h*-Tron turns resistive and a larger share of $I_{bias}$ is diverted towards the main oscillator. This way, the SM of the main oscillator (SM2) can be programmed independently. To program SM1, first, the $I_{gate}$ is kept at zero and $I_{bias}$ is provided to program both SM1 and SM2. To reinstate the SM2 to its original state, the same $I_{bias}$ is provided with the applied $I_{gate}$. Thus, SM1 can be programmed in two steps. Figs. 3(a-f) illustrates the steps for programming the SMs to all four possible combinations of their resistive states.

Figs. 2(g-j) summarizes the spiking dynamics of the proposed neuron circuit. $I_{gate}$ is kept at 0 throughout the whole operation for spike generation. The value of $I_{bias}$ is chosen appropriately to keep both the SNWs (in control and main oscillator) just below their critical current (Fig. 2(g)). In this condition, an equal component of $I_{bias}$ flows through the SNWs of the main oscillator and the control oscillator. When a sufficiently large input current ($I_{input}$) arrives, it increases the total current across the SNW of the main oscillator. This drives the SNW above its $I_c$ and switches it to the resistive state (Fig. 2(h)). The current is redirected to SM2, and the output voltage rises. This reduces the current across the SNW by driving it back to the superconducting state, which in turn relaxes the voltage across the output (Figs. 2(i,j)). Meanwhile, the current across the SNW of the control oscillator increases above $I_c$ and it reinjects current to the loop switching the SNW of the main oscillator. Correspondingly, the output voltage rises again. Thus, two spikes are generated in a single cycle of operation.

Here, for higher $I_{bias}$, the SNW of the main oscillator reaches $I_c$ for a lower injected current in the loop, leading to a quicker oscillation. Therefore, a higher $I_{bias}$ raises the spiking rate. However, a fundamental limit of $I_{bias}$ exists for a successful spike generation process. If $I_{bias} > 2I_c$, both the SNWs will be in the resistive state even in the absence of an applied $I_{in}$. This poses a critical limitation on the reconfigurability of the spiking rate through $I_{bias}$. Moreover, the lowest achievable spiking rate through $I_{bias}$ tuning is primarily governed by the input strength ($I_{in}$) (Fig. 3(e)). For a higher $I_{in}$, a lower $I_{bias}$ can switch the SNW and thereby generate spikes. Thus, higher $I_{in}$ allows a higher range of $I_{bias}$ for modulating the spiking frequency. The inclusion of the dual resistive SM in the design causes an additional tuning knob for the spiking rate. For the same $I_{bias}$ value, four distinct frequencies are achievable for four different combinations of resistive states. In our design, the spike amplitude is determined by the resistance of SM2 and the redirected current across it. Thereby, with two different levels of resistances (LRS and HRS), we achieve two different levels of spike amplitudes. Figs 3(a-d) show the range of achievable spike rate and spike amplitude for four different combinations of SM resistances.

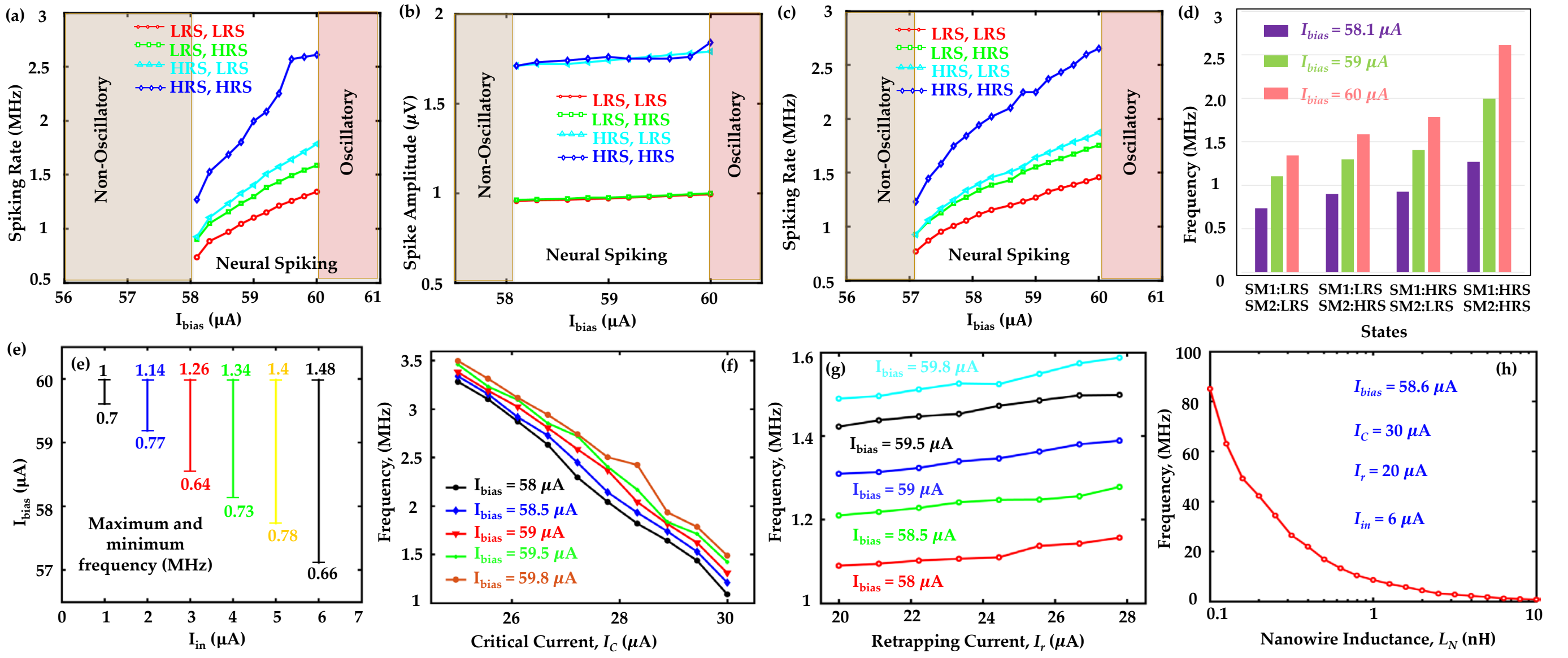

**Fig. 3:** Spiking rate vs $I_{bias}$ for four different combinations of resistive states for **(a)** $I_{in}$ = 4μA and **(b)** $I_{in}$= 6μA. **(c)** Spike amplitude vs $I_{bias}$ for different resistive states. **(d)** Frequency histogram for four different combinations of resistive states for three different values of $I_{bias}$. **(e)** Range of $I_{bias}$ for successful spike generation process for different input strength ($I_{in}$). **(f)** Spiking frequency *vs* critical current ($I_c$) for four different bias current ($I_{bias}$) **(g)** Spiking frequency *vs* retrapping current ($I_r$) for four different $I_{bias}$. **(h)** Spiking frequency *vs* nanowire inductance ($L_N$).

## IV. Design Space and Sensitivity Analysis

We aim to investigate the sensitivity of the spike frequency and power to a wide range of parameters. We intend to vary each parameter independently, without imposing any material constraints, to gain a better understanding of their individual impact. We have considered a range of values that clearly displays the sensitivity of spike generation process on each of these parameters rather choosing a range of values which is quantitatively correct. We begin our analysis by exploring the $I_{bias}$ tunability of the design. The nominal values of the materials and device parameters used in our simulation are summarized in Table I. As shown in Fig. 3(e), for a fixed $I_{in}$, $I_{bias}$ can be varied within a range to une the spike frequencies within a certain range. The maximum $I_{bias}$ for successful spike generation is limited by the $I_c$ of both SNWs. On the other hand, the minimum $I_{bias}$ is controlled by the strength of $I_{in}$. For higher $I_{in}$, minimum applicable $I_{bias}$ reduces and consequently, the range of $I_{bias}$ for successful spike generation process increases linearly (Fig. 3(e)). Below this range, the total currents across the SNWs do not exceed the $I_c$ keeping the output voltage at zero.

We analyze the effect of the material and geometric parameters of both the SNW and SM on the spiking frequency and power. For analyzing the implications of SNW parameters, we vary the current parameters ($I_c$, $I_r$) and nanowire inductance ($L_{NW}$). Additionally, to examine the sensitivity of the SM, we vary their resistance levels and observe their effect on the spike frequency and power. The power is calculated as the average power dissipation from all the bias sources during the spike generation process. While considering the effect of one of these parameters, we keep the others fixed at their nominal values.

*a) Sensitivity to the nanowire parameters:* Here we investigate the influence of material-dictated parameters on the spike generation metrics. $I_c$ and $I_r$ have significant implications on the spike frequency and power because they control the hysteresis window of the SNW. For a fixed $I_r$, higher $I_c$ leads to a broader hysteresis window, resulting in a bigger oscillation period and a lower spiking frequency (Fig. 3(f)). The same reasoning applies in the case of $I_r$ variation. For a fixed $I_c$, a higher $I_r$ implies a narrower hysteresis window, leading to a lower oscillation period and higher spiking frequency (Fig. 3(g)). For both cases, a higher $I_{bias}$ leads to a higher current injection in the loop. This results in a quicker oscillation for both the main and control oscillator block and thus a higher spiking frequency (Figs. 3(f,g)).

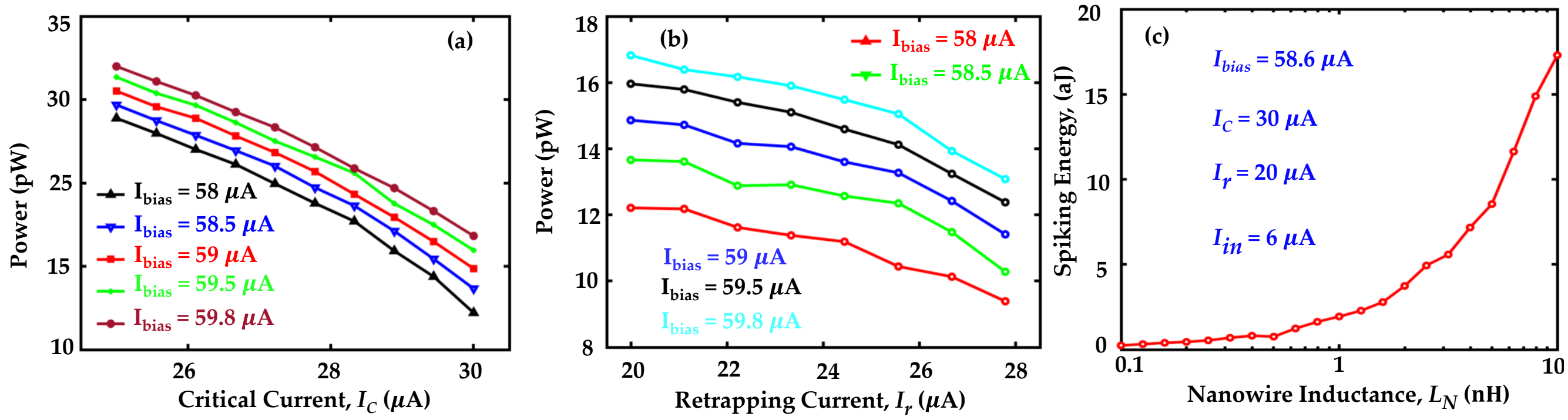

**Fig. 4:** **(a)** Power consumption vs $I_c$ for different bias current ($I_{bias}$). **(b)** Power consumption vs $I_r$ for different $I_{bias}$. **(c)** Spiking energy *vs* $L_N$.

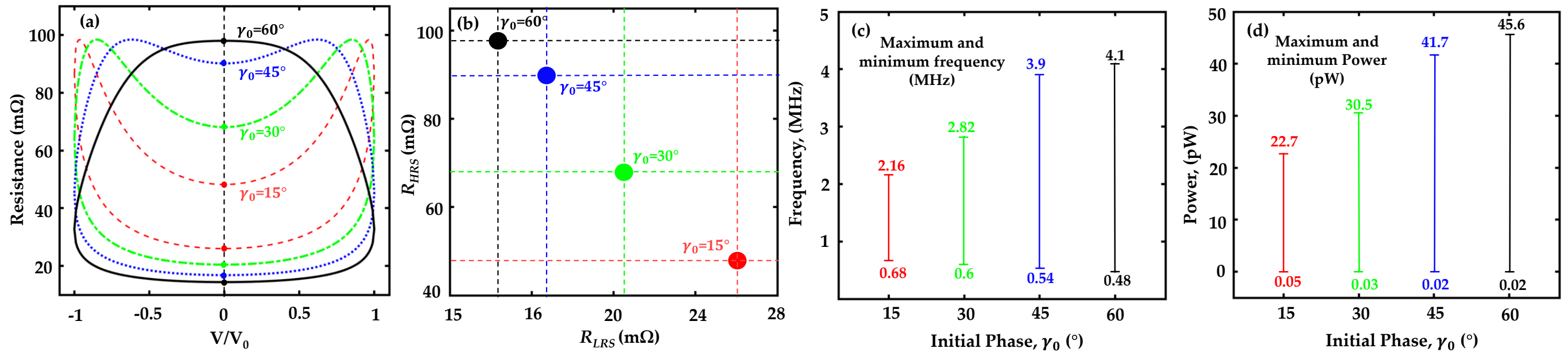

**Fig. 5: (a)** Resistance vs voltage characteristics of SM for four different initial phase parameters ($\gamma_0$ = 15°, 30°, 45°, and 60°). **(b)** $R_{LRS}$ vs $R_{HRS}$ for the four different $\gamma_0$(15°, 30°, 45°, and 60°) considered. Range of **(c)** achievable frequencies, and **(d)** power consumption for four different values of $\gamma_0$.

Both $I_c$ and $I_r$ have important implications on the spiking power dissipation. In this neuron circuit topology, the total power dissipation entirely depends on the current across the SMs during the spike generation process. For a fixed $I_r$, lower $I_c$ leads to a larger share of $I_{bias}$ flowing through the SM, resulting in a higher total power consumption (Fig. 4(a)). Similarly, for a fixed $I_c$, total power consumption increases for a higher $I_r$ as shown in Fig. 4(b). For these two cases, a larger $I_{bias}$ leads to a larger average current across the SMs resulting in a higher total power consumption (Figs. 4(a,b)). On the other hand, the nanowire inductance ($L_{NW}$) has a direct impact on the spiking frequency and power dissipation. The spiking period is primarily dictated by the fall time of the oscillation and the fall time constant ($\tau_2$=$L_{NW}/R_s$) is proportional to $L_{NW}$. Therefore, a higher $L_{NW}$ leads to a higher oscillation period and lower oscillation frequency (Fig. 3(h)). However, as the spiking period increases, the total average power during the spike generation process rises. Hence, a higher inductance results in a higher spiking power (Fig. 4(c)).

*b) Sensitivity on the superconducting memristor:* The phase-dependent conductance of an STJ can be varied for different initial phases ($\gamma_0$) between the two superconducting regions as the LRS and HRS both vary with $\gamma_0$. Here, $\gamma_0$ is the initial difference between the gauge invariant phase ($\gamma$) of the two junctions. It can be tuned externally by inserting the SM structure inside a superconducting loop. The flux produced by the loop can be tuned by varying the current flowing through the loop [34]. Fig.5(a) summarizes the variation in the resistance levels for different $\gamma_0$. As $\gamma_0$ increases from 0 to 90, $R_{LRS}$ decreases and $R_{HRS}$ increases, raising the ratio $R_{HRS}/R_{LRS}$. We have considered four different LRS, HRS pairs for four different $\gamma_0$: 15°, 30°, 60°, 75° (Fig. 5(b)). Figure 5(c) shows the maximum and minimum achievable spiking frequency for four different LRS, HRS pairs. From Figs. 5(a,c), it is evident that the minimum achievable frequency is governed by $R_{LRS}$ of both the SMs. Therefore, as we raise $\gamma_0$, both $R_{LRS}$ and the lowest achievable frequency decrease. Conversely, the maximum achievable frequency is determined by the $R_{HRS}$ of the two SMs. Hence, as shown in Fig. 5(c), with the increase in $\gamma_0$, $R_{HRS}$ and the highest achievable frequency increases. The value of $R_{HRS}$ and $R_{LRS}$ also impacts the maximum and minimum power consumption (Fig. 5(d)). Since the total power consumption is directly proportional to the resistance value ($I^2$·R), an increasing $\gamma_0$leads to the reduction of the lowest achievable power and a rise of the maximum power dissipation, raising the range of power consumption.

**Table II:** Specifications for Monte Carlo Analysis

| Parameters | Mean ($\mu$) | Standard Deviation ($\sigma$) |
|---|---|---|
| Superconducting Nanowire | | |
| Critical current ($I_c$) | 30 $\mu$A | 0.3 $\mu$A |
| Retrapping current ($I_r$) | 20 $\mu$A | 0.2 $\mu$A |
| Nanowire Inductance ($L_{NW}$) | 10 nH | 0.1 nH |
| Superconducting Memristor | | |
| $R_{HRS}$ | 98 mΩ | 1 mΩ |
| $RL_{RS}$ | 14.4 mΩ | 0.15 mΩ |

## V. Variation Analysis

We aim to achieve four distinct levels of spiking frequency for four different combinations of SM resistances. It is of significant importance to ensure that the neuron maintains distinct levels of spiking frequencies and amplitudes under random variation. To examine the effect of variation on the spike frequency and spike amplitude, we consider all the possible sources of variation in our design. The variation in $I_c$, $I_r$, and $L_{NW}$ from SNW are considered. From SM, we consider the variation of $R_{HRS}$ and $R_{LRS}$. Here, we perform a 500-point Monte Carlo variation analysis to observe the variation in the spiking frequency and amplitude. In our analysis, we assume a 3σ-variation of Gaussian distribution for the parameters. Full specifications for our Monte Carlo Analysis are summarized in Table II.

Fig.6 delineates the result of our 500-point Monte Carlo variation analysis for three different values of $I_{bias}$: 58.6 $\mu$A, 59.1 $\mu$A, and 59.6 $\mu$A. Here, four different levels of spiking frequencies are shown, corresponding to the four different combinations of SM resistances. It is evident from the result that our neuron topology maintains a distinct separation between the spiking frequency for (HRS, HRS) and (LRS, HRS) even in the worst case of variation. However, there is a slight overlap between the frequencies for (LRS, HRS) and (HRS, LRS). The separation can be further enhanced by increasing $I_{bias}$ (see Figs. 6(e-l)). Higher $I_{bias}$ results in higher separation between different levels of spiking frequencies. The overlap between the frequencies for (LRS, HRS) and (HRS, LRS) is estimated to be ~14%, ~9%, and ~3% for $I_{bias}$ = 58.6 $\mu$A, 59.1 $\mu$A, and 59.6 $\mu$A respectively. However, there is a distinct separation between the two different spike amplitudes for all three cases proving to be robust against the imposed variations.

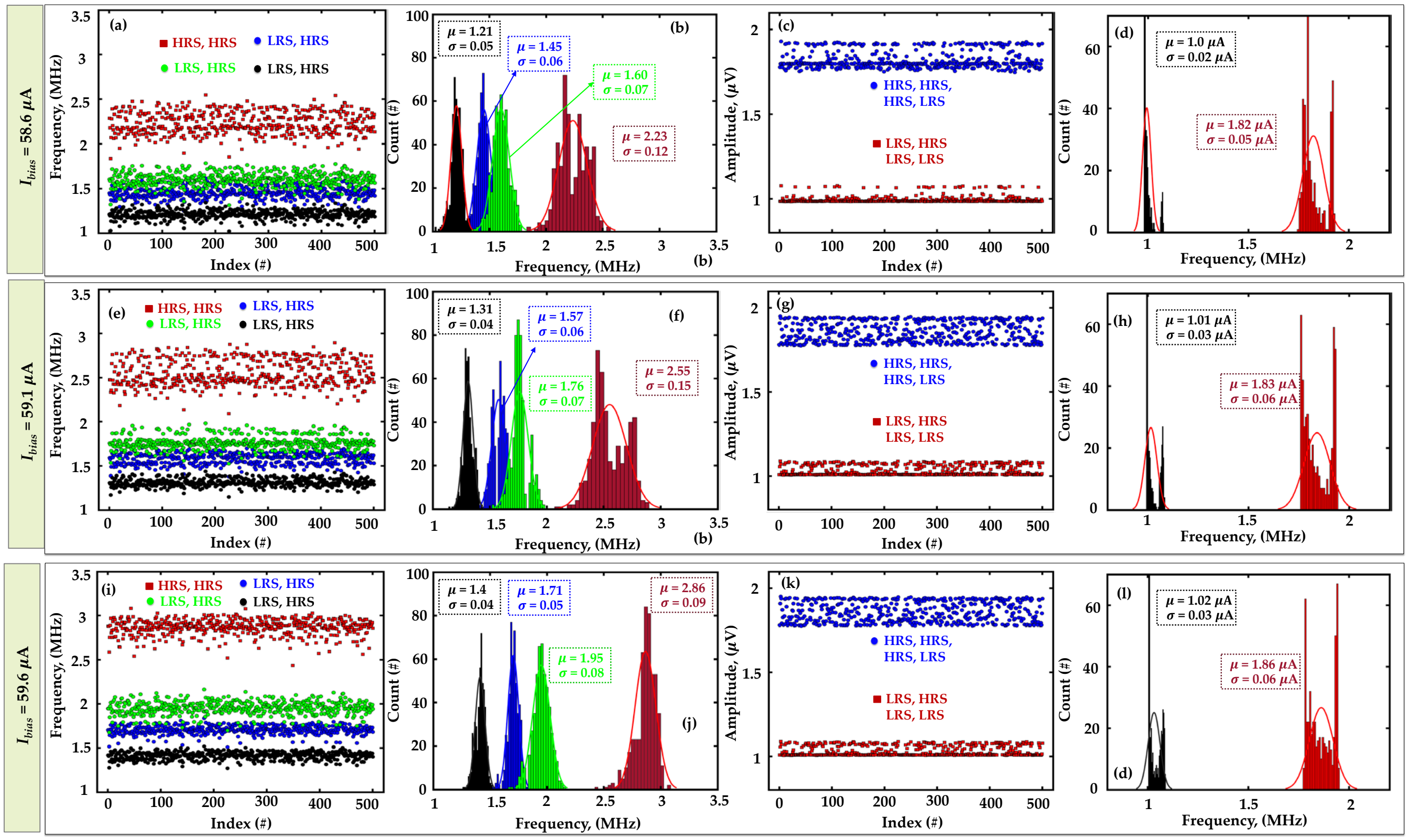

**Fig. 6:** Monte-Carlo variation analysis. **(a)** Scatter plot and **(b)** histogram distribution of four different levels of spike frequencies for $I_{bias}$ = 58.6 $\mu$A. **(c)** Scatter plot and **(d)** histogram distribution of two different levels of spike amplitudes for $I_{bias}$ = 58.6 $\mu$A. **(e)** Scatter plot and **(f)** histogram distribution of four different levels of spike frequencies for $I_{bias}$ = 59.1 $\mu$A. **(g)** Scatter plot and **(h)** histogram distribution of two different levels of spike amplitudes for $I_{bias}$ = 59.1 $\mu$A. **(i)** Scatter plot and **(j)** histogram distribution of four different levels of spike frequencies for $I_{bias}$ = 59.6 $\mu$A. **(k)** Scatter plot and **(l)** histogram distribution of two different levels of spike amplitudes for $I_{bias}$ = 59.6 $\mu$A.

## VI. Conclusion

We performed a comprehensive analysis of the superconducting memristor-based neuron structure to evaluate the dependence of the spike frequency, spike amplitude, and power consumption on all the materials and circuit parameters. From our analysis, we deduced the dependence of spike reconfigurability on the input strength for a successful spike generation process. Our study determines the range of the bias current to dynamically reconfigure the spiking frequency. We also performed an analysis to determine the power consumption of the neuron circuit and examined its dependencies on different circuit materials and circuit parameters to produce valuable insights. In our proposed neuron design, the frequency can be modulated up to ~3.5 times with the aid of bias current and superconducting memristors. The generated insights will be helpful for further research efforts and system-level incorporation of the neuron. Finally, we performed a Monte Carlo variation analysis to further assess the variation robustness of our design. Our result indicates that our neuron design produces distinct levels of spiking frequencies and amplitudes for different combinations of resistive states, and they maintain sufficient separation even in the worst case of variation. Since SM is a JJ-based structure, our design has a lower fanout capability compared to a solely nanowire-based structure. Furthermore, due to the levels of resistances (several tens of millivolts), the amplitude of achieved spikes is at the level of a few microvolts, which further limits the fanout capability of the proposed neuron. We have not estimated the exact fanout capability of our design and it remains as an area for further research.

## Acknowledgment

This material is based in part on research sponsored by Air Force Research Laboratory under agreement FA8750-21-1-1018. The U.S. Government is authorized to reproduce and distribute reprints for Governmental purposes notwithstanding any copyright notation thereon. The views and conclusions contained herein are those of the authors and should not be interpreted as necessarily representing the official policies or endorsements, either expressed or implied, of Air Force Research Laboratory or the U.S. Government.